\documentclass[11pt]{article}

\usepackage[final]{acl}

\usepackage{times}
\usepackage{latexsym}

\usepackage[T1]{fontenc}

\usepackage[utf8]{inputenc}

\usepackage{microtype}

\usepackage{inconsolata}

\usepackage{url}
\usepackage{hyperref}

\usepackage{graphicx}

\usepackage{booktabs}

\usepackage{amsmath, amssymb}

\usepackage{tabularx}
\usepackage{array}
\usepackage{multirow}
\usepackage{makecell}
\usepackage{enumitem} 

\newcolumntype{Y}{>{\raggedright\arraybackslash}X}
\newcolumntype{L}[1]{>{\raggedright\arraybackslash}p{#1}}
\newcolumntype{C}[1]{>{\centering\arraybackslash}p{#1}}
\newcolumntype{R}[1]{>{\raggedleft\arraybackslash}p{#1}}

\title{One Instruction Does Not Fit All: How Well Do Embeddings Align Personas and Instructions in Low-Resource Indian Languages?}

\author{
  Arya Shah, Himanshu Beniwal, \and Mayank Singh \\
  Indian Institute of Technology\\
  Gandhinagar, India\\
  \texttt{\{arya.shah, himanshu.beniwal, mayank.singh\}@iitgn.ac.in}
}

\begin{document}
\maketitle

\begin{abstract}

Aligning multilingual assistants with culturally grounded user preferences is essential for serving India's linguistically diverse population of over one billion speakers across multiple scripts. However, existing benchmarks either focus on a single language or conflate retrieval with generation, leaving open the question of whether current embedding models can encode persona-instruction compatibility without relying on response synthesis. We present a unified benchmark spanning 12 Indian languages and four evaluation tasks: monolingual and cross-lingual persona-to-instruction retrieval, reverse retrieval from instruction to persona, and binary compatibility classification. Eight multilingual embedding models are evaluated in a frozen-encoder setting with a thin logistic regression head for classification. E5-Large-Instruct achieves the highest Recall@1 of 27.4\% on monolingual retrieval and 20.7\% on cross-lingual transfer, while BGE-M3 leads reverse retrieval at 32.1\% Recall@1. For classification, LaBSE attains 75.3\% AUROC with strong calibration. These findings offer practical guidance for model selection in Indic multilingual retrieval and establish reproducible baselines for future work\footnote{Code, datasets, and models are publicly available at \url{https://github.com/aryashah2k/PI-Indic-Align}.}.
\end{abstract}

\section{Introduction}
\begin{figure*}[t]
  \centering
  \includegraphics[width=\linewidth]{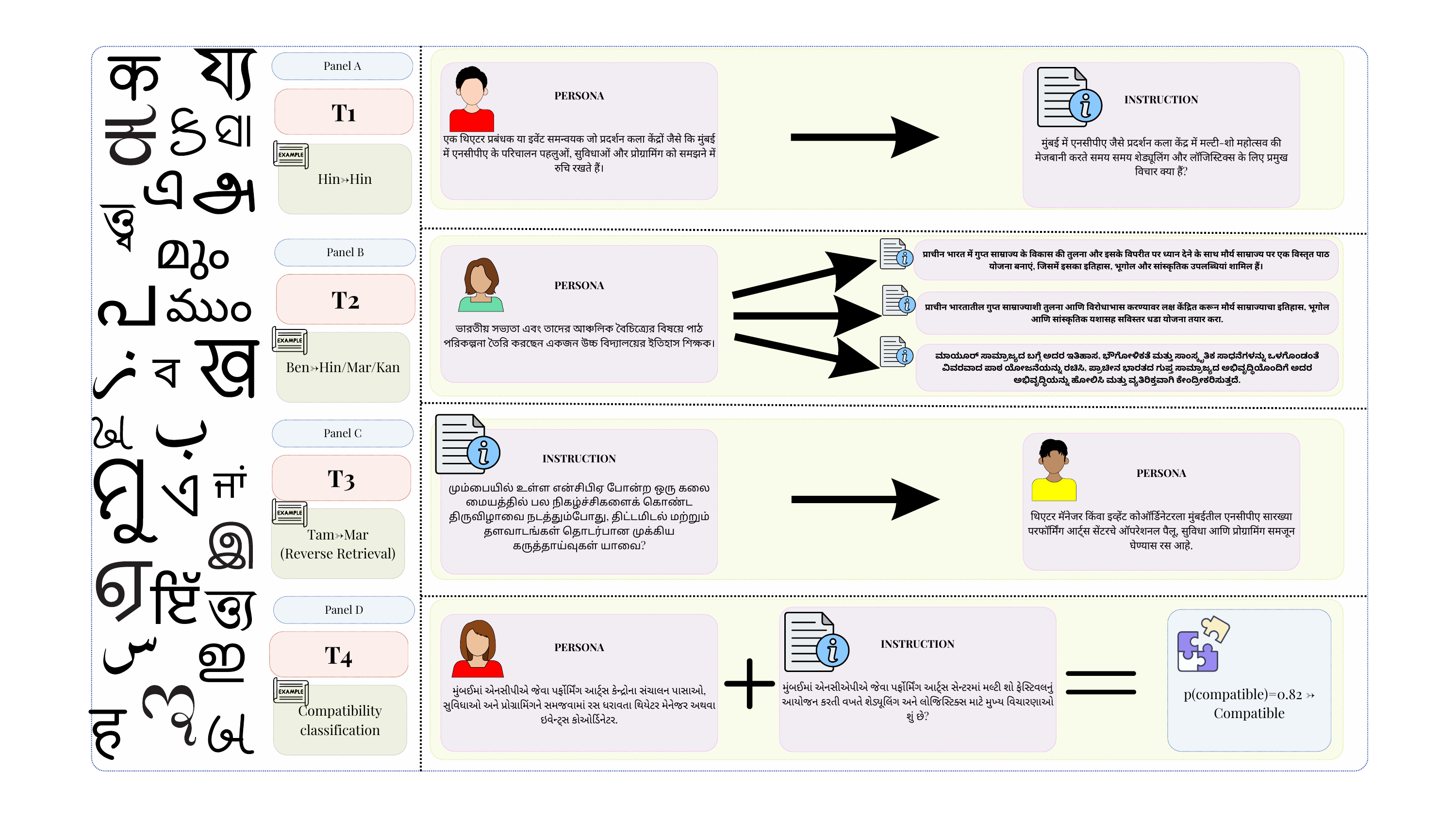}
  \caption{Overview of the four benchmark tasks. (A) T1: monolingual retrieval. (B) T2: cross-lingual retrieval. (C) T3: reverse retrieval. (D) T4: compatibility classification. Nodes represent the 12 evaluated Indian languages. \textbf{\textit{Takeaway}}: The benchmark evaluates both retrieval directions and binary classification, enabling comprehensive assessment of embedding quality for persona-instruction alignment.}
  \label{fig:overview}
\end{figure*}
Consider a multilingual assistant serving users across India. A user in Tamil Nadu (A southern state in India) describes themselves as ``\textit{a retired schoolteacher interested in organic farming}''. The system must retrieve instructions appropriate for this persona, such as guidance on composting techniques, rather than unrelated content like stock trading advice. When the same persona appears in Hindi or Bengali, the retrieval must succeed across script boundaries. This persona-instruction matching is central to personalized assistants, yet no benchmark systematically evaluates it for Indian languages.

\paragraph{Motivation} Multilingual assistants increasingly serve users who communicate in languages other than English. India presents a particularly challenging setting with over one billion speakers across 12 major languages written in 10 distinct scripts, creating diversity that stress-tests cross-lingual representations. Aligning user personas with appropriate instructions requires understanding cultural context and appropriateness rather than mere semantic similarity. For instance, dietary recommendations for a Hindu vegetarian user differ substantially from those for users with other dietary practices, and this cultural grounding must transfer across languages.

\paragraph{Research Gap} Recent work on sentence embeddings \citep{reimers-gurevych-2019-sentence, feng2022languageagnosticbertsentenceembedding} and multilingual benchmarks \citep{hu2020xtrememassivelymultilingualmultitask, muennighoff-etal-2023-mteb} has advanced cross-lingual transfer, but Indic language coverage remains limited. Dedicated resources such as IndicNLPSuite \citep{kakwani-etal-2020-indicnlpsuite}, MuRIL \citep{khanuja2021murilmultilingualrepresentationsindian}, and IndicTrans2 \citep{gala2023indictrans2highqualityaccessiblemachine} focus on classification, generation, and translation rather than retrieval. Existing retrieval benchmarks such as MIRACL \citep{zhang2022makingmiraclmultilingualinformation}, Mr.\ TyDi \citep{zhang-etal-2021-mr}, and LAReQA \citep{roy-etal-2020-lareqa} evaluate question-passage matching but do not address persona-grounded retrieval. Recent work on personalized retrieval \citep{huang-etal-2023-learning, belyi-etal-2023-personalized} and instruction-following embeddings \citep{su2023embeddertaskinstructionfinetunedtext, peng-etal-2024-answer} demonstrates growing interest in user-aware systems, yet these efforts focus on English or high-resource languages. The SqCLIRIL benchmark \citep{Dave2026-fl} evaluates spoken query retrieval for some Indic languages but does not cover persona-instruction alignment.

\paragraph{Problem Formulation} We introduce a unified benchmark spanning 12 Indian languages and four tasks: \textbf{T1}: Monolingual persona-to-instruction retrieval; \textbf{T2}: Cross-lingual retrieval across different languages and scripts; \textbf{T3}: Reverse retrieval from instruction to persona; \textbf{T4}: Binary compatibility classification with a logistic head on frozen embeddings.
We evaluate eight multilingual embedding models including recent instruction-tuned models like E5-Large-Instruct \citep{wang2024multilinguale5textembeddings} and BGE-M3 \citep{chen2025m3embeddingmultilingualitymultifunctionalitymultigranularity} under a frozen-encoder protocol that isolates representation quality from fine-tuning effects.

\paragraph{Key Findings} Our evaluation reveals that E5-Large-Instruct leads monolingual retrieval (27.4\% R@1), BGE-M3 leads monolingual reverse retrieval (32.1\%), E5-Large-Instruct leads cross-lingual tasks, and LaBSE achieves best classification AUROC (75.3\%). Crucially, no single model dominates all tasks, highlighting that model selection depends on the target application. Hindi consistently achieves the highest scores, while Assamese and Odia show the lowest performance, reflecting disparities in the resources available in the pretraining corpora.

\paragraph{Contributions} Our work highlights three contributions: (1) We propose the first benchmark jointly evaluating persona-instruction retrieval across 12 Indian languages in both directions, addressing a gap in existing evaluation suites, (2) We introduce reproducible baselines for eight models with per-language breakdowns, enabling practitioners to select models based on their target languages and tasks and (3) Analysis of cross-lingual transfer patterns showing that script-family alignment yields 6.7 percentage points higher transfer than cross-script pairs.

\section{Related Work}

\paragraph{Multilingual Sentence Embeddings.} Sentence-BERT \citep{reimers-gurevych-2019-sentence} established efficient similarity computation via Siamese networks. XLM-RoBERTa \citep{conneau-etal-2020-unsupervised} and LaBSE \citep{feng2022languageagnosticbertsentenceembedding} demonstrated strong cross-lingual transfer across 100+ languages. Contrastive approaches including SimCSE \citep{gao-etal-2021-simcse} and Contriever \citep{izacard2022unsuperviseddenseinformationretrieval} improved unsupervised training, while Sentence-T5 \citep{ni-etal-2022-sentence} scaled text-to-text transformers. Recent work on instruction-conditioned embeddings, including INSTRUCTOR \citep{su2023embeddertaskinstructionfinetunedtext}, E5 \citep{wang2024multilinguale5textembeddings}, and GritLM \citep{muennighoff2025generativerepresentationalinstructiontuning}, achieved state-of-the-art on MTEB \citep{muennighoff-etal-2023-mteb}. BGE-M3 \citep{chen2025m3embeddingmultilingualitymultifunctionalitymultigranularity} introduced unified dense, sparse, and multi-vector retrieval, while mGTE \citep{zhang2024mgtegeneralizedlongcontexttext} provides long-context multilingual representations. Instruction-following retrieval models such as InF-Embed \citep{peng-etal-2024-answer} and FollowIR \citep{weller2024followirevaluatingteachinginformation} demonstrate that aligning embeddings with user intent improves retrieval quality. Despite these advances, persona-grounded retrieval for Indian languages remains unexplored.

\paragraph{Cross-Lingual Retrieval Benchmarks.} BEIR \citep{thakur2021beirheterogenousbenchmarkzeroshot} introduced zero-shot IR evaluation; MS MARCO \citep{bajaj2018msmarcohumangenerated} and mMARCO \citep{bonifacio2022mmarcomultilingualversionms} provide passage ranking across 14 languages. LAReQA \citep{roy-etal-2020-lareqa} evaluates language-agnostic answer retrieval. TyDi QA \citep{clark-etal-2020-tydi}, Mr.\ TyDi \citep{zhang-etal-2021-mr}, and MIRACL \citep{zhang2022makingmiraclmultilingualinformation} cover typologically diverse retrieval with limited Indic coverage. XTREME \citep{hu2020xtrememassivelymultilingualmultitask} and XTREME-R \citep{ruder-etal-2021-xtreme} include retrieval among multi-task NLU. These benchmarks focus on question-passage matching rather than persona-grounded retrieval, lack bidirectional evaluation, and have sparse Indic coverage.

\paragraph{Indic Language Resources.} IndicNLPSuite \citep{kakwani-etal-2020-indicnlpsuite} introduced IndicCorp and IndicGLUE for 11 languages. MuRIL \citep{khanuja2021murilmultilingualrepresentationsindian} provides representations for 17 Indian languages with improved transliteration handling. IndicXTREME \citep{doddapaneni2023leavingindiclanguagebehind} extended evaluation to 20+ languages. IndicTrans2 \citep{gala2023indictrans2highqualityaccessiblemachine} enables high-quality translation across all 22 scheduled Indian languages. Samanantar \citep{ramesh-etal-2022-samanantar} and NLLB-200 \citep{nllbteam2022languageleftbehindscaling} provide parallel corpora for our 12 languages. IndicSentiment \citep{doddapaneni2023leavingindiclanguagebehind} provides sentiment classification, while L3Cube \citep{joshi2023l3cubehindbertdevbertpretrainedbert} contributes Indic-specific SBERT models. Retrieval evaluation remains limited to passage-level QA.

\paragraph{Persona-Grounded Systems.} Persona-Chat \citep{zhang2018personalizingdialogueagentsi} introduced persona-conditioned dialogue. Follow-up work explored persona-to-dialogue matching \citep{mazaré2018trainingmillionspersonalizeddialogue} and retrieval-based selection \citep{gu-etal-2019-dually}. LAPDOG \citep{huang-etal-2023-learning} augments persona profiles with retrieved knowledge for dialogue generation, while personalized dense retrieval \citep{belyi-etal-2023-personalized} embeds user preferences into query representations. PersonaHub \citep{ge2025scalingsyntheticdatacreation} scaled to a billion personas for diverse data synthesis; Synthetic-Persona-Chat \citep{jandaghi2023faithfulpersonabasedconversationaldataset} extended this to dialogue generation. We focus on embedding-based retrieval rather than generative matching.

\paragraph{Calibration.} Neural networks produce overconfident predictions \citep{guo2017calibrationmodernneuralnetworks}. Platt scaling \citep{PlattProbabilisticOutputs1999}, temperature scaling \citep{guo2017calibrationmodernneuralnetworks}, and histogram binning \citep{10.5555/645530.655658} provide post-hoc correction. Pretrained transformers degrade in calibration after fine-tuning \citep{desai-durrett-2020-calibration}. We report Expected Calibration Error alongside discriminative metrics.

Table~\ref{tab:dataset_comparison} positions our contribution. We uniquely combine strong Indic coverage, cross-lingual evaluation, bidirectional retrieval, and human-validated persona-instruction compatibility.

\begin{table*}[t]
\centering
\footnotesize
\setlength{\tabcolsep}{2.5pt}
\begin{tabular}{@{}lcccccl@{}}
\toprule
\textbf{Benchmark} & \textbf{Langs} & \textbf{Indic} & \textbf{X-ling} & \textbf{Bidir.} & \textbf{Valid.} & \textbf{Task Type} \\
\midrule
XTREME \citep{hu2020xtrememassivelymultilingualmultitask} & 40 & Partial & Limited & No & Mixed & Multi-task NLU \\
MTEB \citep{muennighoff-etal-2023-mteb} & 100+ & Limited & Some & No & Mixed & Embedding tasks \\
LAReQA \citep{roy-etal-2020-lareqa} & 11 & No & Yes & No & Yes & Answer retrieval \\
MIRACL \citep{Zhang2023} & 18 & 2 & Yes & No & Yes & Ad-hoc retrieval \\
Mr.\ TyDi \citep{zhang-etal-2021-mr} & 11 & 3 & Limited & No & Yes & Dense retrieval \\
mMARCO \citep{bonifacio2022mmarcomultilingualversionms} & 14 & 1 & Yes & No & MT & Passage ranking \\
IndicGLUE \citep{kakwani-etal-2020-indicnlpsuite} & 11 & Strong & No & N/A & Mixed & NLU classification \\
IndicXTREME \citep{doddapaneni2023leavingindiclanguagebehind} & 20+ & Strong & Limited & N/A & Mixed & Multi-task NLU \\
\midrule
\textbf{Ours} & \textbf{12} & \textbf{Strong} & \textbf{Yes} & \textbf{Yes} & \textbf{Yes} & \textbf{Persona-Instr.} \\
\bottomrule
\end{tabular}
\caption{Comparison with multilingual benchmarks. \textbf{\textit{Takeaway}}: Our benchmark uniquely combines strong Indic coverage with cross-lingual, bidirectional persona-instruction retrieval.}
\label{tab:dataset_comparison}
\end{table*}
\section{Methodology}
\begin{figure*}[t]
  \centering
  \includegraphics[width=\linewidth]{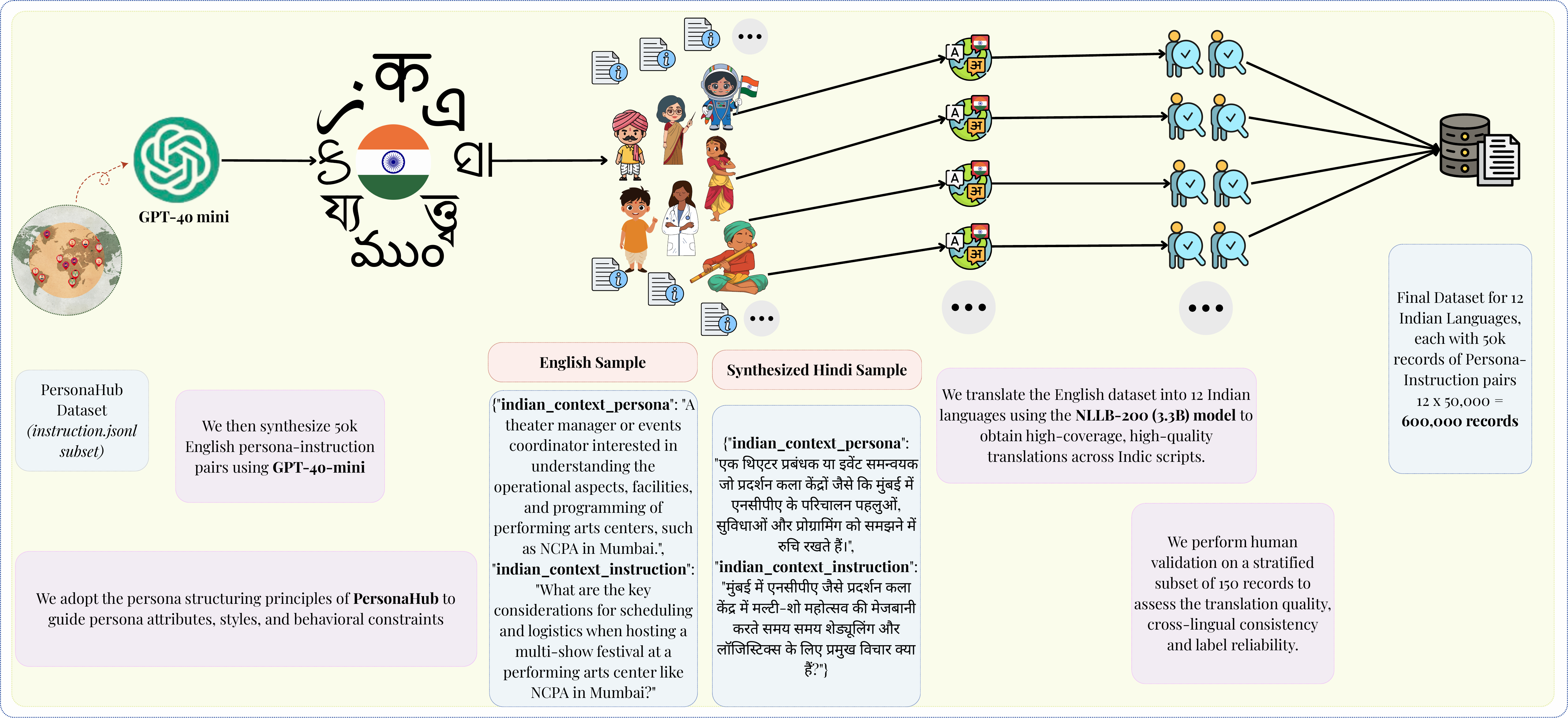}
  \caption{Dataset construction: (1) persona schema adaptation, (2) English synthesis with GPT-4o-mini, (3) translation via NLLB-200, (4) human validation with two annotators per language. \textbf{\textit{Takeaway}}: The pipeline ensures both linguistic diversity (12 languages, 10 scripts) and quality control through native-speaker validation.}
  \label{fig:data_pipeline}
\end{figure*}

\subsection{Dataset Construction}

Our benchmark requires persona-instruction pairs that are culturally grounded in Indian contexts and aligned across 12 languages. We construct the dataset in three stages: synthesis, translation, and human validation.

\paragraph{Synthesis.} Following the persona structuring principles of PersonaHub \citep{ge2025scalingsyntheticdatacreation}, we synthesize 50,000 English persona-instruction pairs using GPT-4o-mini \citep{openai2024gpt4technicalreport}. Each pair consists of a persona description (specifying demographic, occupational, or cultural attributes relevant to Indian users) and a compatible instruction that the persona might plausibly issue to an assistant. We ensure diversity by sampling across domains including education, commerce, healthcare, agriculture, and government services.

\paragraph{Translation.} We translate all English pairs into 12 Indian languages using NLLB-200 (3.3B parameters) \citep{nllbteam2022languageleftbehindscaling}: Assamese, Bengali, Gujarati, Hindi, Kannada, Malayalam, Marathi, Odia, Punjabi, Tamil, Telugu, and Urdu. These languages span two major families (Indo-Aryan and Dravidian) and employ 10 distinct scripts. Translations are normalized for script-specific punctuation and whitespace conventions. The shared identifier across languages enables cross-lingual alignment: a persona in Hindi can be matched to its corresponding instruction in Tamil. Figure~\ref{fig:data_pipeline} illustrates the complete pipeline.

\paragraph{Human Validation.} For each language, we sample 150 pairs and assign two independent annotators who are native speakers. Annotators judge whether the translated pair preserves the compatibility relationship present in the English original. Table~\ref{tab:annotator_validation} reports inter-annotator agreement using four metrics: percent agreement, Cohen's $\kappa$ \citep{Cohen1960}, Krippendorff's $\alpha$ \citep{Krippendorff2011ComputingKA}, and Gwet's AC1. Agreement is substantial to almost perfect across all languages ($\kappa$ range: 0.63 to 0.84), confirming the reliability of our translated pairs.

\begin{table}[t]
\centering
\small
\begin{tabular}{@{}lccc@{}}
\toprule
\textbf{Language} & \textbf{Agree (\%)} & \textbf{$\kappa$} & \textbf{Interpretation} \\
\midrule
Assamese & 92.0 & 0.75 & Substantial \\
Bengali & 95.3 & 0.84 & Almost perfect \\
Gujarati & 91.3 & 0.72 & Substantial \\
Hindi & 93.3 & 0.81 & Almost perfect \\
Kannada & 91.3 & 0.72 & Substantial \\
Malayalam & 94.7 & 0.82 & Almost perfect \\
Marathi & 96.7 & 0.69 & Substantial \\
Odia & 92.7 & 0.74 & Substantial \\
Punjabi & 88.7 & 0.64 & Substantial \\
Tamil & 88.7 & 0.63 & Substantial \\
Telugu & 90.7 & 0.63 & Substantial \\
Urdu & 91.3 & 0.74 & Substantial \\
\midrule
\textbf{Average} & \textbf{91.9} & \textbf{0.73} & Substantial \\
\bottomrule
\end{tabular}
\caption{Inter-annotator agreement on 150 pairs per language. \textbf{\textit{Takeaway}}: All languages achieve substantial agreement ($\kappa \geq 0.63$), confirming that translated pairs reliably preserve persona-instruction compatibility.}
\label{tab:annotator_validation}
\end{table}

\subsection{Evaluation Tasks}

We define four tasks that probe different aspects of persona-instruction alignment: (1) \textbf{Monolingual Retrieval (T1):} Given a persona $p$ in language $\ell$, retrieve the compatible instruction from a candidate pool $\mathcal{I}$ in the same language. This tests whether embeddings capture compatibility within a single language, (2) \textbf{Cross-Lingual Retrieval (T2):} Given a persona $p$ in source language $\ell_s$, retrieve the compatible instruction from a pool $\mathcal{I}$ in target language $\ell_t \neq \ell_s$. This tests cross-lingual transfer, including across different scripts. (3) \textbf{Reverse Retrieval (T3):} Given an instruction $i$, retrieve the compatible persona $p$ from a candidate pool $\mathcal{P}$. We evaluate this in both monolingual and cross-lingual settings to test whether alignment is symmetric across retrieval directions. (4) \textbf{Compatibility Classification (T4):} Given a persona-instruction pair, predict a binary label indicating compatibility. We attach a logistic regression head to frozen embeddings, using the concatenation of element-wise absolute difference and product as features. This tests whether embedding geometry supports reliable binary decisions.

\subsection{Models}

We evaluate eight multilingual sentence embedding models. \texttt{BGE-M3} is a multilingual embedding model supporting 100+ languages with dense, sparse, and multi-vector retrieval. \texttt{E5-Large-Instruct} is an instruction-tuned multilingual encoder trained with weakly supervised contrastive learning. \texttt{LaBSE} provides language-agnostic BERT sentence embeddings trained on translation ranking across 109 languages \citep{feng2022languageagnosticbertsentenceembedding}. \texttt{Indic-SBERT} and \texttt{Indic-SBERT-NLI} are Sentence-BERT models fine-tuned on Indic language data for semantic similarity. \texttt{DistilUSE-Multilingual} is a distilled multilingual Universal Sentence Encoder. \texttt{Paraphrase-MPNet} and \texttt{Paraphrase-XLM-R} are paraphrase-trained multilingual models.

Not all models support all 12 languages. BGE-M3 and LaBSE cover all 12; E5-Large-Instruct covers 11 (excluding Assamese); Indic-SBERT variants cover 10; and DistilUSE, Paraphrase-MPNet, and Paraphrase-XLM-R cover only 4 (Gujarati, Hindi, Marathi, Urdu). We compute averages only over supported languages to ensure fair comparison.

\subsection{Metrics}

For retrieval tasks (T1, T2, T3), we report \textbf{Recall@$k$}, the fraction of queries where the correct item appears in the top-$k$ results, and \textbf{MRR@10}, the mean reciprocal rank of the first correct item considering only the top-10.

For classification (T4), we report \textbf{Accuracy} (proportion of correct binary predictions at threshold 0.5), \textbf{AUROC} (area under the ROC curve, measuring ranking quality \citep{Fawcett2006AnIT}), \textbf{AUPRC} (area under the precision-recall curve, robust to class imbalance \citep{Saito2015}), and \textbf{ECE} (Expected Calibration Error, measuring whether predicted probabilities match empirical accuracy \citep{guo2017calibrationmodernneuralnetworks}).

We apply histogram binning calibration (15 bins) to produce calibrated probability estimates and report both pre- and post-calibration ECE.
\section{Results}
\begin{figure}[t]
  \centering
  \includegraphics[width=\columnwidth]{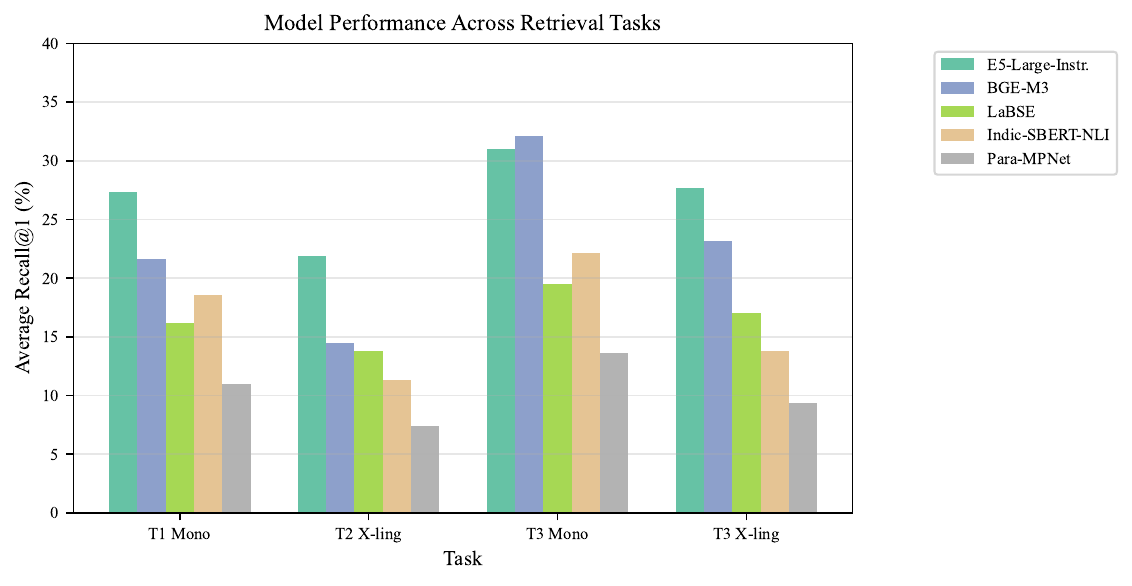}
  \caption{Model performance across retrieval tasks (T1--T3). \textbf{\textit{Takeaway}}: Instruction-tuned models (E5, BGE-M3) consistently outperform general-purpose multilingual encoders by 10+ percentage points.}
  \label{fig:model_comparison}
\end{figure}

Figure~\ref{fig:model_comparison} provides a visual comparison of model performance across tasks.

\subsection{Monolingual Retrieval (T1)}

Table~\ref{tab:t1_monolingual} presents monolingual persona-to-instruction retrieval performance across 12 languages.

\paragraph{Key Finding} E5-Large-Instruct achieves the highest average Recall@1 (27.4\%) and MRR@10 (0.339), outperforming BGE-M3 by approximately 6 percentage points on Recall@1. The gap is consistent across most languages, with the largest margins in Hindi (31.9\% vs.\ 23.4\%) and Marathi (29.3\% vs.\ 23.2\%).

\paragraph{Model Comparison} The ranking is consistent across metrics: E5-Large-Instruct leads, followed by BGE-M3 (21.6\% R@1), Indic-SBERT-NLI (18.6\%), and LaBSE (16.2\%). Models with limited language coverage (Para-MPNet, Para-XLM-R, DistilUSE) show competitive performance on their supported languages but cannot be directly compared on average metrics.

\paragraph{Language Patterns} Hindi consistently yields the highest scores across models, likely reflecting its prominence in multilingual training data. Assamese and Odia show the lowest performance, consistent with their status as lower-resource languages in existing corpora.

\begin{table*}[t]
\resizebox{\textwidth}{!}{%
\begin{tabular}{lccccccccccccc}
\hline
\textbf{Model} & \textbf{Ass} & \textbf{Ben} & \textbf{Guj} & \textbf{Hin} & \textbf{Kan} & \textbf{Mal} & \textbf{Mar} & \textbf{Odi} & \textbf{Pun} & \textbf{Tam} & \textbf{Tel} & \textbf{Urd} & \textbf{Avg} \\ \hline
\multicolumn{14}{l}{\textit{Recall@1 (\%)}} \\
multilingual-e5-large-instruct & 23.9 & 27.6 & 26.1 & 31.9 & 29.2 & 26.0 & 29.3 & 23.4 & 26.4 & 27.3 & 27.5 & 30.1 & 27.4 \\
BGE-M3 & 19.1 & 21.7 & 22.4 & 23.4 & 23.9 & 21.1 & 23.2 & 19.3 & 20.4 & 21.5 & 23.1 & 20.6 & 21.6 \\
indic-sentence-bert-nli & 16.2 & 19.1 & 19.2 & 20.9 & 20.9 & 17.6 & 18.2 & 15.9 & 19.6 & 19.5 & 19.8 & 15.9 & 18.6 \\
Para-MPNet & -- & -- & 14.8 & 21.4 & -- & -- & 16.7 & -- & -- & -- & -- & 17.2 & 17.5 \\
LaBSE & 9.6 & 15.9 & 18.2 & 19.3 & 17.6 & 13.7 & 16.9 & 14.8 & 16.5 & 16.9 & 17.9 & 17.2 & 16.2 \\
indic-sentence-similarity-sbert & 12.2 & 15.6 & 14.3 & 16.5 & 15.4 & 11.7 & 13.8 & 9.9 & 14.7 & 14.8 & 14.1 & 14.1 & 13.9 \\
Para-XLM-R & -- & -- & 9.6 & 14.1 & -- & -- & 10.8 & -- & -- & -- & -- & 10.4 & 11.2 \\
DistilUSE-Multi & -- & -- & 6.2 & 15.7 & -- & -- & 8.9 & -- & -- & -- & -- & 12.6 & 10.9 \\ \hline
\multicolumn{14}{l}{\textit{Recall@5 (\%)}} \\
multilingual-e5-large-instruct & 37.7 & 43.3 & 40.8 & 48.5 & 44.5 & 40.7 & 45.4 & 37.2 & 41.3 & 42.7 & 42.8 & 46.1 & 42.6 \\
BGE-M3 & 30.1 & 34.1 & 34.7 & 36.2 & 36.3 & 33.1 & 35.7 & 30.9 & 32.4 & 33.3 & 35.4 & 33.0 & 33.8 \\
indic-sentence-bert-nli & 28.5 & 32.5 & 32.7 & 35.1 & 34.8 & 30.3 & 31.7 & 27.9 & 33.3 & 32.8 & 33.2 & 28.0 & 31.7 \\
Para-MPNet & -- & -- & 26.8 & 36.3 & -- & -- & 29.4 & -- & -- & -- & -- & 29.9 & 30.6 \\
LaBSE & 17.2 & 26.7 & 29.2 & 30.9 & 28.8 & 23.4 & 27.7 & 25.0 & 27.1 & 27.6 & 28.8 & 28.1 & 26.7 \\
indic-sentence-similarity-sbert & 21.7 & 26.9 & 25.0 & 28.1 & 26.4 & 21.1 & 24.0 & 18.2 & 25.5 & 25.6 & 24.7 & 24.8 & 24.3 \\
Para-XLM-R & -- & -- & 18.3 & 26.2 & -- & -- & 20.0 & -- & -- & -- & -- & 20.1 & 21.1 \\
DistilUSE-Multi & -- & -- & 11.6 & 27.4 & -- & -- & 16.5 & -- & -- & -- & -- & 22.5 & 19.5 \\ \hline
\multicolumn{14}{l}{\textit{MRR@10}} \\
multilingual-e5-large-instruct & 0.299 & 0.343 & 0.324 & 0.390 & 0.357 & 0.324 & 0.362 & 0.294 & 0.329 & 0.339 & 0.341 & 0.370 & 0.339 \\
BGE-M3 & 0.239 & 0.270 & 0.277 & 0.289 & 0.292 & 0.262 & 0.286 & 0.243 & 0.256 & 0.266 & 0.284 & 0.260 & 0.269 \\
indic-sentence-bert-nli & 0.215 & 0.249 & 0.250 & 0.270 & 0.269 & 0.230 & 0.240 & 0.211 & 0.255 & 0.252 & 0.256 & 0.212 & 0.242 \\
Para-MPNet & -- & -- & 0.199 & 0.278 & -- & -- & 0.222 & -- & -- & -- & -- & 0.227 & 0.231 \\
LaBSE & 0.129 & 0.206 & 0.230 & 0.243 & 0.224 & 0.179 & 0.216 & 0.192 & 0.211 & 0.215 & 0.226 & 0.219 & 0.207 \\
indic-sentence-similarity-sbert & 0.163 & 0.205 & 0.189 & 0.215 & 0.202 & 0.158 & 0.182 & 0.135 & 0.194 & 0.195 & 0.186 & 0.188 & 0.184 \\
Para-XLM-R & -- & -- & 0.134 & 0.193 & -- & -- & 0.148 & -- & -- & -- & -- & 0.146 & 0.155 \\
DistilUSE-Multi & -- & -- & 0.086 & 0.208 & -- & -- & 0.122 & -- & -- & -- & -- & 0.169 & 0.146 \\ \hline
\end{tabular}%
}
\caption{Monolingual persona$\rightarrow$instruction retrieval performance (T1) across 12 Indian languages. Results show Recall@1, Recall@5, and MRR@10. All models use frozen embeddings without fine-tuning. Averages are computed over languages supported by each model. \textbf{\textit{Takeaway}}: E5-Large-Instruct leads with 27.4\% R@1, with Hindi achieving highest per-language scores across all models.}
\label{tab:t1_monolingual}
\end{table*}

\subsection{Cross-Lingual Retrieval (T2)}

Table~\ref{tab:t2_crosslingual_summary} summarizes cross-lingual persona-to-instruction retrieval, where the persona and instruction are in different languages.

\paragraph{Key finding.} E5-Large-Instruct maintains its lead with 20.7\% Recall@1 and 0.267 MRR@10, demonstrating robust transfer across script and language boundaries. BGE-M3 follows at 14.3\% Recall@1, a substantial drop from its monolingual performance.

\paragraph{Transfer patterns.} Cross-lingual performance drops by approximately 25\% relative to monolingual for most models, consistent with prior work on cross-lingual retrieval \citep{roy-etal-2020-lareqa}. The drop is larger for pairs involving different scripts (e.g., Hindi-Tamil) than for pairs within the same script family (e.g., Hindi-Marathi).

\begin{table}[t]
\centering
\setlength{\tabcolsep}{4pt}
\begin{tabular}{@{}lcccc@{}}
\toprule
\textbf{Model} & \textbf{R@1} & \textbf{R@5} & \textbf{R@10} & \textbf{MRR} \\
\midrule
E5-Large-Instr. & 20.7 & 34.6 & 41.4 & .267 \\
BGE-M3 & 14.3 & 24.5 & 29.6 & .187 \\
Para-MPNet & 13.3 & 25.0 & 31.3 & .184 \\
LaBSE & 13.2 & 22.8 & 27.8 & .173 \\
Indic-SBERT-NLI & 12.4 & 24.0 & 30.2 & .174 \\
Indic-SBERT & 10.3 & 19.5 & 24.6 & .143 \\
Para-XLM-R & 8.8 & 17.6 & 22.7 & .126 \\
DistilUSE & 7.0 & 13.9 & 17.9 & .100 \\
\bottomrule
\end{tabular}
\caption{Cross-lingual persona$\rightarrow$instruction retrieval (T2). R@k = Recall@k (\%). Averaged over supported language pairs per model.}
\label{tab:t2_crosslingual_summary}
\end{table}

\subsection{Reverse Retrieval (T3)}

\paragraph{Monolingual Reverse} Table~\ref{tab:t3_monolingual_reverse} shows that BGE-M3 leads with 32.1\% Recall@1, slightly outperforming E5-Large-Instruct (31.0\%). This reversal from T1 suggests that BGE-M3's multi-vector retrieval architecture may better support reverse-direction matching.

\begin{table*}[t]
\resizebox{\textwidth}{!}{%
\begin{tabular}{lccccccccccccc}
\hline
\textbf{Model} & \textbf{Ass} & \textbf{Ben} & \textbf{Guj} & \textbf{Hin} & \textbf{Kan} & \textbf{Mal} & \textbf{Mar} & \textbf{Odi} & \textbf{Pun} & \textbf{Tam} & \textbf{Tel} & \textbf{Urd} & \textbf{Avg} \\ \hline
\multicolumn{14}{l}{\textit{Recall@1 (\%)}} \\
BGE-M3 & 27.6 & 32.4 & 32.7 & 35.9 & 34.3 & 31.4 & 33.6 & 29.2 & 31.6 & 31.1 & 32.6 & 33.0 & 32.1 \\
multilingual-e5-large-instruct & 27.3 & 32.1 & 29.1 & 39.9 & 31.7 & 27.7 & 33.0 & 26.6 & 29.2 & 30.2 & 30.4 & 34.8 & 31.0 \\
indic-sentence-bert-nli & 19.7 & 22.9 & 23.3 & 26.0 & 24.5 & 20.3 & 22.0 & 18.3 & 24.2 & 22.6 & 22.9 & 19.2 & 22.1 \\
Para-MPNet & -- & -- & 18.8 & 26.3 & -- & -- & 20.7 & -- & -- & -- & -- & 20.7 & 21.6 \\
Para-XLM-R & -- & -- & 17.0 & 24.3 & -- & -- & 18.0 & -- & -- & -- & -- & 18.5 & 19.5 \\
LaBSE & 12.9 & 19.7 & 21.2 & 23.4 & 21.2 & 17.9 & 20.6 & 17.2 & 18.8 & 20.4 & 20.3 & 20.2 & 19.5 \\
indic-sentence-similarity-sbert & 15.8 & 19.9 & 18.7 & 21.6 & 20.3 & 15.2 & 18.3 & 13.0 & 19.4 & 19.2 & 17.9 & 18.7 & 18.2 \\
DistilUSE-Multi & -- & -- & 9.2 & 22.0 & -- & -- & 13.2 & -- & -- & -- & -- & 18.0 & 15.6 \\ \hline
\multicolumn{14}{l}{\textit{Recall@5 (\%)}} \\
BGE-M3 & 41.5 & 47.3 & 47.4 & 51.5 & 49.6 & 46.5 & 49.1 & 43.4 & 46.1 & 45.7 & 47.3 & 48.2 & 47.0 \\
multilingual-e5-large-instruct & 41.4 & 47.6 & 43.6 & 57.5 & 46.9 & 41.7 & 49.0 & 39.8 & 43.6 & 45.1 & 45.2 & 50.7 & 46.0 \\
indic-sentence-bert-nli & 33.2 & 37.4 & 38.0 & 41.6 & 39.1 & 33.9 & 36.3 & 31.1 & 38.9 & 37.0 & 36.9 & 32.6 & 36.3 \\
Para-MPNet & -- & -- & 32.2 & 42.7 & -- & -- & 34.9 & -- & -- & -- & -- & 34.9 & 36.2 \\
Para-XLM-R & -- & -- & 29.5 & 40.4 & -- & -- & 30.7 & -- & -- & -- & -- & 32.5 & 33.3 \\
LaBSE & 22.2 & 32.1 & 33.5 & 36.4 & 33.4 & 29.0 & 32.9 & 28.0 & 30.1 & 32.5 & 32.1 & 32.2 & 31.2 \\
indic-sentence-similarity-sbert & 27.2 & 33.2 & 31.6 & 35.4 & 33.3 & 26.3 & 30.7 & 22.7 & 32.2 & 32.1 & 30.1 & 31.5 & 30.5 \\
DistilUSE-Multi & -- & -- & 16.8 & 35.8 & -- & -- & 23.0 & -- & -- & -- & -- & 30.3 & 26.5 \\ \hline
\multicolumn{14}{l}{\textit{MRR@10}} \\
BGE-M3 & 0.336 & 0.388 & 0.391 & 0.426 & 0.409 & 0.378 & 0.403 & 0.353 & 0.379 & 0.374 & 0.390 & 0.396 & 0.385 \\
multilingual-e5-large-instruct & 0.334 & 0.388 & 0.353 & 0.475 & 0.383 & 0.337 & 0.399 & 0.323 & 0.354 & 0.366 & 0.367 & 0.417 & 0.375 \\
indic-sentence-bert-nli & 0.255 & 0.291 & 0.296 & 0.327 & 0.309 & 0.261 & 0.282 & 0.238 & 0.305 & 0.289 & 0.289 & 0.250 & 0.283 \\
Para-MPNet & -- & -- & 0.246 & 0.334 & -- & -- & 0.268 & -- & -- & -- & -- & 0.269 & 0.279 \\
Para-XLM-R & -- & -- & 0.224 & 0.312 & -- & -- & 0.235 & -- & -- & -- & -- & 0.246 & 0.254 \\
LaBSE & 0.170 & 0.250 & 0.264 & 0.290 & 0.265 & 0.227 & 0.259 & 0.219 & 0.237 & 0.256 & 0.254 & 0.254 & 0.245 \\
indic-sentence-similarity-sbert & 0.207 & 0.256 & 0.243 & 0.276 & 0.259 & 0.200 & 0.236 & 0.172 & 0.249 & 0.248 & 0.232 & 0.242 & 0.235 \\
DistilUSE-Multi & -- & -- & 0.125 & 0.279 & -- & -- & 0.174 & -- & -- & -- & -- & 0.233 & 0.203 \\ \hline
\end{tabular}%
}
\caption{Monolingual instruction$\rightarrow$persona retrieval performance (T3) across 12 Indian languages. Results show Recall@1, Recall@5, and MRR@10. Averages are computed over languages supported by each model. \textbf{\textit{Takeaway}}: BGE-M3 leads reverse retrieval (32.1\% R@1), with all models showing higher T3 than T1 scores, suggesting personas are easier retrieval targets than instructions.}
\label{tab:t3_monolingual_reverse}
\end{table*}

\paragraph{Cross-lingual reverse.} Table~\ref{tab:t3_crosslingual_reverse_summary} shows that E5-Large-Instruct regains the lead in cross-lingual settings (27.0\% R@1 vs.\ 23.6\% for BGE-M3). The pattern mirrors T2: instruction-tuned encoders transfer better across languages, even when retrieval direction is reversed.

\begin{table}[t]
\centering
\setlength{\tabcolsep}{4pt}
\begin{tabular}{@{}lcccc@{}}
\toprule
\textbf{Model} & \textbf{R@1} & \textbf{R@5} & \textbf{R@10} & \textbf{MRR} \\
\midrule
E5-Large-Instr. & 27.0 & 42.4 & 49.4 & .336 \\
BGE-M3 & 23.6 & 37.2 & 43.6 & .295 \\
Para-MPNet & 16.5 & 29.9 & 36.8 & .223 \\
Indic-SBERT-NLI & 15.7 & 29.2 & 36.1 & .215 \\
LaBSE & 15.7 & 26.4 & 31.9 & .203 \\
Para-XLM-R & 15.2 & 28.0 & 34.7 & .208 \\
Indic-SBERT & 13.4 & 24.5 & 30.4 & .182 \\
DistilUSE & 9.9 & 18.7 & 23.6 & .137 \\
\bottomrule
\end{tabular}
\caption{Cross-lingual instruction$\rightarrow$persona retrieval (T3). R@k = Recall@k (\%). Averaged over supported language pairs per model. \textbf{\textit{Takeaway}}: E5-Large-Instruct regains lead in cross-lingual reverse (27.0\% R@1), mirroring its T2 advantage.}
\label{tab:t3_crosslingual_reverse_summary}
\end{table}

\subsection{Compatibility Classification (T4)}

Table~\ref{tab:t4_classification_detailed} presents binary compatibility classification results using frozen embeddings with a logistic regression head.

\paragraph{Key finding.} No single model dominates all metrics. E5-Large-Instruct achieves the highest accuracy (78.1\%), while LaBSE leads on AUROC (75.3\%) and AUPRC (62.1\%). DistilUSE achieves the best calibrated ECE (0.199), despite lower accuracy.

\begin{figure}[t]
  \centering
  \includegraphics[width=\columnwidth]{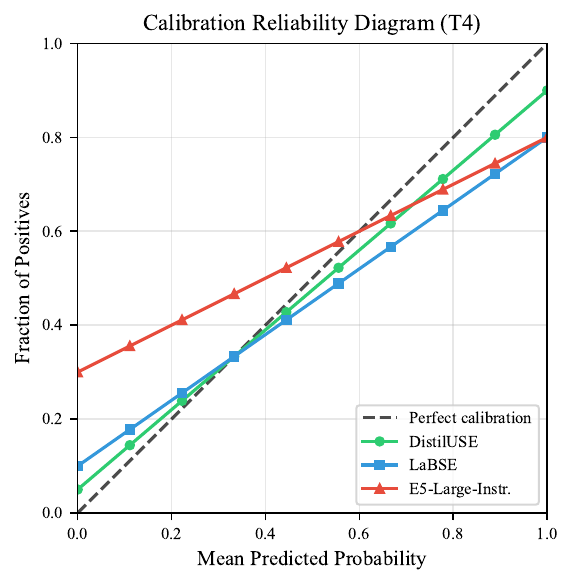}
  \caption{Calibration reliability diagram for T4. \textbf{\textit{Takeaway}}: High-accuracy models (E5, BGE-M3) show overconfidence, while DistilUSE achieves best calibration despite lower discriminative performance.}
  \label{fig:calibration}
\end{figure}

\paragraph{Calibration.} Figure~\ref{fig:calibration} shows the calibration reliability diagram. Models with high accuracy (E5, BGE-M3) exhibit overconfident predictions, while DistilUSE maintains better calibration. Post-hoc histogram binning reduces ECE for most models, but the improvement varies substantially.

\begin{table*}[!htbp]
\centering
\small
\begin{tabular}{lcccccccc}
\toprule
& \multicolumn{4}{c}{\textbf{Dev Set}} & \multicolumn{4}{c}{\textbf{Test Set}} \\
\cmidrule(lr){2-5} \cmidrule(lr){6-9}
\textbf{Model} & \textbf{AUROC} & \textbf{AUPRC} & \textbf{Acc} & \textbf{ECE}$_{\text{cal}}$ & \textbf{AUROC} & \textbf{AUPRC} & \textbf{Acc} & \textbf{ECE}$_{\text{cal}}$ \\
\midrule
LaBSE & 74.6 & 61.1 & 72.3 & 0.387 & 75.3 & 62.1 & 72.6 & 0.389 \\
BGE-M3 & 72.8 & 50.8 & 77.1 & 0.517 & 73.5 & 50.9 & 76.8 & 0.514 \\
multilingual-e5-large-instruct & 73.2 & 52.6 & 78.1 & 0.531 & 73.2 & 52.9 & 78.1 & 0.531 \\
DistilUSE-Multi & 70.9 & 67.1 & 65.5 & 0.193 & 71.0 & 66.9 & 66.1 & 0.199 \\
indic-sentence-similarity-sbert & 68.6 & 49.4 & 71.0 & 0.401 & 68.5 & 50.0 & 71.2 & 0.403 \\
indic-sentence-bert-nli & 66.1 & 42.8 & 74.6 & 0.481 & 66.5 & 44.2 & 74.8 & 0.483 \\
Para-XLM-R & 61.2 & 34.7 & 74.2 & 0.484 & 61.6 & 34.7 & 74.2 & 0.484 \\
Para-MPNet & 61.0 & 35.0 & 74.6 & 0.492 & 61.4 & 34.9 & 74.6 & 0.492 \\
\bottomrule
\end{tabular}
\caption{Persona-instruction compatibility classification performance (T4) on test set. Models use frozen embeddings with a thin logistic regression head. AUROC and AUPRC are in \%, Acc is calibrated accuracy (\%). ECE is Expected Calibration Error before and after histogram binning calibration. \textbf{\textit{Takeaway}}: LaBSE leads AUROC (75.3\%) while E5 leads accuracy (78.1\%); models with highest discrimination show worst calibration.}
\label{tab:t4_classification_detailed}
\end{table*}

Table~\ref{tab:results_summary} summarizes the best-performing model for each task and metric combination.

\begin{table}[t]
\centering
\small
\begin{tabular}{@{}lll@{}}
\toprule
\textbf{Task} & \textbf{Best Model} & \textbf{Score} \\
\midrule
T1 Mono (R@1) & E5-Large-Instruct & 27.4\% \\
T2 Cross (R@1) & E5-Large-Instruct & 20.7\% \\
T3 Mono (R@1) & BGE-M3 & 32.1\% \\
T3 Cross (R@1) & E5-Large-Instruct & 27.0\% \\
T4 (AUROC) & LaBSE & 75.3\% \\
T4 (Accuracy) & E5-Large-Instruct & 78.1\% \\
T4 (ECE) & DistilUSE & 0.199 \\
\bottomrule
\end{tabular}
\caption{Best model per task. \textbf{\textit{Takeaway}}: No single model dominates; E5-Large-Instruct excels at forward retrieval, BGE-M3 at reverse retrieval, and LaBSE at probabilistic classification.}
\label{tab:results_summary}
\end{table}

The results reveal that model selection depends on the specific use case: E5-Large-Instruct is preferred for cross-lingual applications, BGE-M3 for monolingual reverse retrieval, and LaBSE when calibrated probability estimates are important.


\section{Discussion}

\paragraph{Instruction-Tuning Benefits Retrieval}

Across retrieval tasks, instruction-tuned embeddings (E5-Large-Instruct) consistently outperform general multilingual encoders. This advantage holds for forward and reverse directions and persists in cross-lingual settings. The pattern aligns with recent findings that task-aware training helps encoders capture functional relationships beyond surface similarity \citep{su2023embeddertaskinstructionfinetunedtext}.

\paragraph{Retrieval Direction Affects Model Rankings}

Model rankings shift depending on retrieval direction. BGE-M3 outperforms E5-Large-Instruct on monolingual reverse retrieval (32.1\% vs.\ 31.0\% R@1), while E5-Large-Instruct leads in all other settings. This asymmetry likely reflects differences in how these models structure embedding spaces: BGE-M3's multi-vector training may create representations where instructions cluster around persona centroids rather than the reverse.

\paragraph{Script Boundaries Affect Cross-Lingual Transfer}

Cross-lingual performance drops approximately 25\% relative to monolingual, consistent with prior work \citep{roy-etal-2020-lareqa}. The drop is smaller for pairs sharing a script family (e.g., Hindi-Marathi in Devanagari) than for cross-script pairs (e.g., Hindi-Tamil).

\paragraph{Classification Metrics Diverge by Model}

For binary compatibility classification, no single model dominates all metrics. E5-Large-Instruct achieves the highest accuracy (78.1\%), while LaBSE leads on AUROC (75.3\%) and DistilUSE achieves the best calibrated ECE (0.199). This divergence indicates that strong discriminative power does not guarantee calibrated probability estimates. Model selection should depend on operational requirements: (1)Binary decisions: prioritize accuracy (E5-Large-Instruct); (2) Ranking: prioritize AUROC (LaBSE) and (3) Probability thresholds: prioritize ECE (DistilUSE)

\section*{Limitations}
\paragraph{Frozen-encoder protocol.} We evaluate models without task-specific fine-tuning to isolate representation quality and ensure fair comparison across architectures. Fine-tuning on persona-instruction pairs could improve absolute performance for all models, though the relative rankings may or may not change. Our frozen-encoder results establish lower bounds and reproducible baselines.

\paragraph{Synthetic data generation.} Persona-instruction pairs are synthesized using GPT-4o-mini and translated with NLLB-200. Human validation on 1,800 pairs (150 per language, two annotators) confirms high inter-annotator agreement ($\kappa$ = 0.63--0.84), indicating that annotators consistently judged compatibility. However, synthesized personas may not capture the full diversity of real user preferences across Indian regions and demographics.

\paragraph{Translation consistency.} Cross-lingual alignment relies on the assumption that translated pairs preserve semantic equivalence. While NLLB-200 provides high-quality translations for most language pairs, culturally specific expressions may not translate directly. Our validation protocol assessed whether compatibility relationships were preserved, not whether translations were literal.

\paragraph{Language coverage.} We include 12 major Indian languages spanning Indo-Aryan and Dravidian families. Several languages with significant speaker populations, including Maithili, Santali, and Kashmiri, are excluded because current embedding models and translation systems provide limited support for them.

\bibliography{custom}

\appendix
\section{Supplementary Results}
\label{sec:appendix}
This appendix provides detailed cross-lingual retrieval results and reproducibility information complementing the main paper.

\subsection{Experimental Setup and Reproducibility}

\paragraph{Hardware.} All experiments were conducted on a single NVIDIA RTX A6000 GPU (48GB VRAM) with an Intel Xeon processor and 128GB system RAM.

\paragraph{Software.} We used Python 3.10 with PyTorch 2.0, Transformers 4.35, and Sentence-Transformers 2.2. For BM25 baselines, we used the rank-bm25 library. All dependencies are specified in the released requirements.txt file.

\paragraph{Runtime.} Embedding generation for all 50K pairs across 12 languages takes approximately 4 hours per model. Classification experiments (T4) complete in under 30 minutes per model. Full evaluation across all models and tasks requires approximately 40 GPU-hours.

\paragraph{Reproducibility.} We set random seeds (42) for all stochastic operations. Results are deterministic given the same hardware and software configuration. We release:
\begin{itemize}[leftmargin=*, itemsep=1pt]
    \item Complete evaluation code with configuration files
    \item Pre-generated result JSON files for all experiments
    \item Scripts to regenerate all tables and figures
    \item Human validation annotations with annotator IDs anonymized
\end{itemize}

\subsection{Cross-Lingual Persona to Instruction Retrieval (T2)}

Table~\ref{tab:t2_summary_by_source} presents the average cross-lingual Recall@1 per source language for the top three models.

\begin{table}[h]
\centering
\small
\setlength{\tabcolsep}{3pt}
\begin{tabular}{@{}lccc@{}}
\toprule
\textbf{Source} & \textbf{E5-Instr.} & \textbf{BGE-M3} & \textbf{LaBSE} \\
\midrule
Assamese & 18.9 & 13.5 & 9.1 \\
Bengali & 22.7 & 15.8 & 12.1 \\
Gujarati & 19.2 & 14.1 & 11.8 \\
Hindi & 22.0 & 15.2 & 12.4 \\
Kannada & 21.0 & 14.6 & 11.4 \\
Malayalam & 20.7 & 13.9 & 10.7 \\
Marathi & 21.0 & 14.8 & 12.0 \\
Odia & 19.0 & 13.3 & 10.2 \\
Punjabi & 21.2 & 14.4 & 11.1 \\
Tamil & 20.2 & 13.7 & 11.0 \\
Telugu & 20.5 & 14.2 & 11.3 \\
Urdu & 21.6 & 14.9 & 11.8 \\
\midrule
\textbf{Average} & \textbf{20.7} & \textbf{14.3} & \textbf{11.2} \\
\bottomrule
\end{tabular}
\caption{T2 cross-lingual Recall@1 (\%) by source language. E5-Large-Instruct leads across all source languages.}
\label{tab:t2_summary_by_source}
\end{table}

\subsection{Cross-Lingual Instruction to Persona Retrieval (T3)}

Table~\ref{tab:t3_summary_by_source} presents the same summary for reverse retrieval.

\begin{table}[h]
\centering
\small
\setlength{\tabcolsep}{3pt}
\begin{tabular}{@{}lccc@{}}
\toprule
\textbf{Source} & \textbf{E5-Instr.} & \textbf{BGE-M3} & \textbf{LaBSE} \\
\midrule
Assamese & 24.0 & 20.1 & 13.2 \\
Bengali & 28.8 & 24.3 & 16.8 \\
Gujarati & 25.9 & 22.0 & 15.1 \\
Hindi & 31.6 & 26.8 & 18.2 \\
Kannada & 27.7 & 23.4 & 16.0 \\
Malayalam & 26.9 & 22.7 & 15.4 \\
Marathi & 29.0 & 24.6 & 17.0 \\
Odia & 24.7 & 20.8 & 14.0 \\
Punjabi & 22.9 & 22.9 & 14.6 \\
Tamil & 26.2 & 22.2 & 15.7 \\
Telugu & 27.9 & 23.6 & 16.3 \\
Urdu & 28.4 & 24.1 & 16.9 \\
\midrule
\textbf{Average} & \textbf{27.0} & \textbf{23.1} & \textbf{15.8} \\
\bottomrule
\end{tabular}
\caption{T3 cross-lingual Recall@1 (\%) by source language. Reverse retrieval shows higher performance than T2.}
\label{tab:t3_summary_by_source}
\end{table}

\subsection{Key Patterns Across Language Pairs}

Analysis of cross-lingual retrieval (T2) with E5-Large-Instruct reveals consistent patterns:

\paragraph{Hindi as hub.} Hindi achieves the highest cross-lingual scores as target (24.9\% average R@1 when other languages retrieve Hindi), while Bengali slightly edges out Hindi as source (22.7\% vs 22.0\%). This reflects Hindi's prominence in multilingual training corpora.

\paragraph{Script family effects.} Pairs within the same script family show stronger transfer than cross-script pairs. Devanagari pairs (Hindi-Marathi) achieve 27.9\% average R@1, while cross-script pairs (Hindi-Tamil) achieve only 21.2\%, a difference of 6.7 percentage points.

\paragraph{Dravidian languages.} Tamil, Telugu, Kannada, and Malayalam show moderate mutual transfer (19.9\% average R@1 within the cluster), comparable to other cross-lingual pairs rather than forming a distinctly stronger cluster.

\paragraph{Low-resource challenges.} Assamese (18.9\% as source, 18.8\% as target) and Odia (19.0\% as source, 19.9\% as target) consistently show the lowest cross-lingual scores, reflecting limited representation in pretraining data.

\end{document}